\documentclass[runningheads,notitlepage]{llncs}
\usepackage{graphicx}
\usepackage{comment}
\usepackage{epsfig}
\usepackage{graphicx}
\usepackage{multirow}
\usepackage{xcolor}
\usepackage{soul}
\usepackage{enumerate}
\usepackage{amsmath,amssymb} %
\usepackage{xspace}
\usepackage{cleveref}
\usepackage{subcaption}
\usepackage{wrapfig}

\renewcommand{\paragraph}[1]{\vspace{.5em} \noindent{\textbf{#1}}}

\makeatletter
\DeclareRobustCommand\onedot{\futurelet\@let@token\@onedot}
\def\@onedot{\ifx\@let@token.\else.\null\fi\xspace}

\def\eg{\emph{e.g}\onedot} 
\def\ie{\emph{i.e}\onedot}

\makeatother

\newcommand{\x}{\mathbf{x}}
\newcommand{\D}{{\mathcal{D}}}
\newcommand{\Df}{{\mathcal{D}_f}}
\newcommand{\Dr}{{\mathcal{D}_r}}
\newcommand{\Dtest}{{\mathcal{D}_{\text{Test}}}}
\newcommand{\G}{{\nabla f_{0}(\D)}}
\newcommand{\Gr}{{\nabla f_{0}(\Dr)}}
\newcommand{\Gf}{{\nabla f_{0}(\Df)}}
\newcommand{\Trr}{\Theta_{rr}}
\newcommand{\Tff}{\Theta_{ff}}
\newcommand{\Trf}{\Theta_{rf}}
\newcommand{\lin}{\text{lin}}
\definecolor{applegreen}{rgb}{0.55, 0.71, 0.0}
\definecolor{asparagus}{rgb}{0.53, 0.66, 0.42}
\definecolor{ao}{rgb}{0.0, 0.5, 0.0}

\newcommand{\utext}[2]{\underbrace{#1}_{\text{#2}}}

\newcommand{\E}{\mathbb{E}}

\newcommand{\KL}[2]{\operatorname{KL}\big({\textstyle{#1}\,\|\,{#2}}\big)}

\DeclareMathOperator{\tr}{tr}

\begin{document}
\pagestyle{headings}
\mainmatter
\def\ECCVSubNumber{6746}  %

\title{Forgetting Outside the Box:\\
Scrubbing Deep Networks of Information Accessible from Input-Output Observations}

\titlerunning{Forgetting outside the box}
\authorrunning{Golatkar, Achille, Soatto}

\author{Aditya Golatkar \and
Alessandro Achille \and
Stefano Soatto}
\institute{Department of Computer Science \\
University of California, Los Angeles\\
\email{\{aditya29,achille,soatto\}@cs.ucla.edu}}
\maketitle

\begin{abstract}
We describe a procedure for removing dependency on a cohort of training data from a trained deep network %
that improves upon and generalizes previous methods to different readout functions, and can be extended to ensure forgetting in the final activations of the network. We introduce a new bound on how much information can be extracted per query about the forgotten cohort from a black-box network for which only the input-output behavior is observed. The proposed forgetting procedure has a deterministic part derived from the differential equations of a linearized version of the model, and a stochastic part that ensures information destruction by adding noise tailored to the geometry of the loss landscape. We exploit the connections between the final activations and weight dynamics of a DNN inspired by Neural Tangent Kernels to compute the information in the final activations.
\keywords{Forgetting, Data Removal, Neural Tangent Kernel, Information Theory}
\end{abstract}

\begin{figure}[t]
    \centering
    \hspace{1.5cm}
    \includegraphics[height=2.7cm]{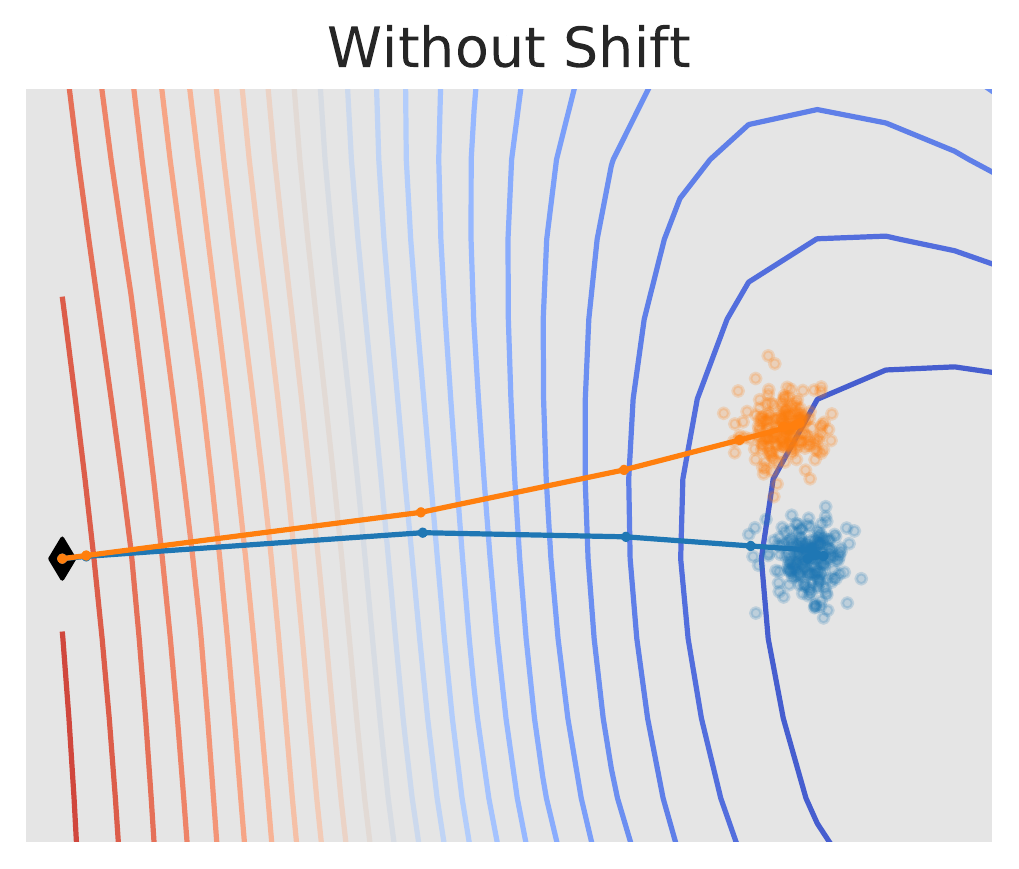}
    \hspace{.2cm}
    \includegraphics[height=2.7cm]{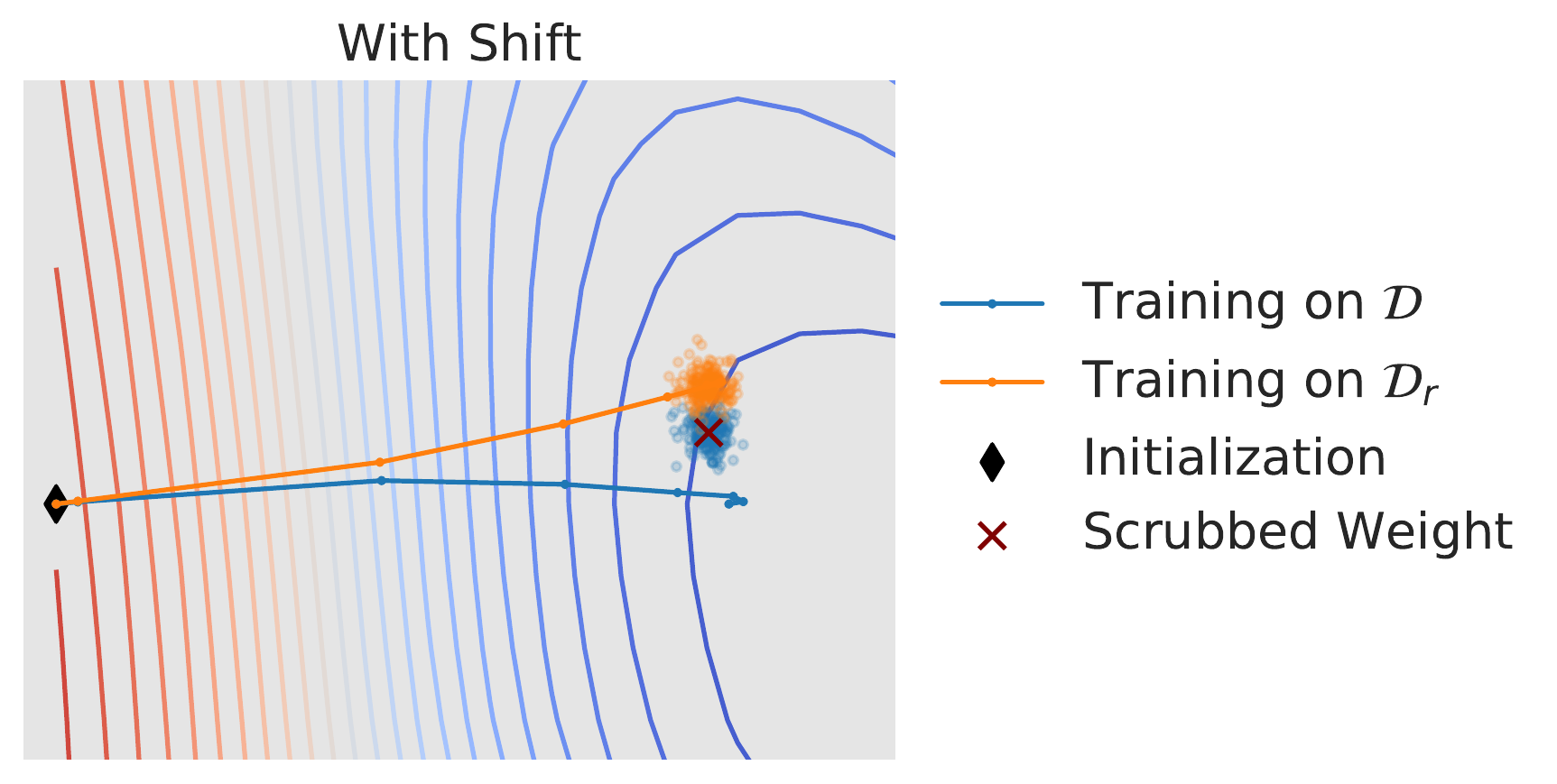}
    \caption{\textbf{Scrubbing procedure:} PCA-projection of training paths on $\D$ (blue), $\Dr$ (orange) and the weights after scrubbing, using \textbf{(Left)} The Fisher method of \cite{golatkar2019eternal}, and \textbf{(Right)} the proposed scrubbing method. Our proposed scrubbing procedure (red cross) moves the model towards $w(\Dr)$, which reduces the amount of noise (point cloud) that needs to be added to achieve forgetting.}
    \label{fig:pca-plot}%
\end{figure}

\section{Introduction}
\label{sec:introduction}

We study the problem of removing information pertaining to a given set of data points from the weights of a trained network. After removal, a potential attacker should not be able to recover information about the forgotten cohort. We consider both the cases in which the attacker has full access to the weights of the trained model, and the less-studied case where the attacker can only query the model by observing some input data and the corresponding output, for instance through a web Application Programming Interface (API).
We show that we can quantify the maximum amount of information that an attacker can extract from observing inputs and outputs (black-box attack), as well as from direct knowledge of the weights (white-box), and propose tailored procedures for removing such information from the trained model in {\em one shot}. That is, assuming the model has been obtained by fine-tuning a pre-trained generic backbone, we compute a single perturbation of the weights that, in one go, can erase information about a cohort to be forgotten in such a way that an attacker cannot access it.

More formally, let a dataset $\D$ be partitioned into a subset  $\Df$ to be forgotten, and its complement $\Dr$ to be retained, $\D = \Df \sqcup \Dr$. A (possibly stochastic) training algorithm $A$ takes $\Df$ and $\Dr$ and outputs a weight vector $w \in \mathbb{R}^{p}$:
\[A(\Df, \Dr) \to w.\]
Assuming an attacker knows the training algorithm $A$ (\eg, stochastic gradient descent, or SGD), the weights $w$, and the retainable data $\Dr$, she can exploit their relationship to recover information about $\Df$, at least for state-of-the-art deep neural networks (DNNs) using a ``readout'' function \cite{golatkar2019eternal}, that is, a function $R(w)$ that an attacker can apply on the weights of a DNN to extract information about the dataset.
For example, an attacker may measure the confidence of the classifier on an input, or measure the time that it takes to fine-tune the network on a given subset of samples to decide whether that input or set was used to train the classifier (membership attack). We discuss  additional examples in \Cref{sec:readout}. Ideally, the forgetting procedure should be robust to different choices of readout functions.
Recent work \cite{golatkar2019eternal,guo2019certified}, introduces a ``scrubbing procedure'' (forgetting procedure or data deletion/removal) $S_\Df(w) : \mathbb{R}^{p} \rightarrow \mathbb{R}^{p}$, that attempts to remove information about $\Df$ from the weights, {\em i.e.},
\[A(\Df, \Dr) \to w \to S_\Df(w)\]
with an upper-bound on the amount of information about $\Df$ that can be extracted after the forgetting procedure, provided the attack has access to the scrubbed weights $S_\Df(w)$, a process called ``white-box attack.''

Bounding the information that can be extracted from a white-box attack is often complex and may be overly restrictive: Deep networks have large sets of equivalent solutions -- a ``null space'' --  that would give the same activations on all test samples. Changes in $\Df$ may change the position of the weights in the null space. Hence, the position of the weight in the null-space, even if irrelevant for the input-output behavior, may be exploited to recover information about $\Df$.

This suggests that the study of forgetting should be approached from the perspective of the {\em final activations}, rather than the weights, since there could be infinitely many different models that produce the same input-output behavior, and we are interested in preventing attacks that affect the behavior of the network, rather than the specific solution to which the training process converged.  More precisely, denote by $f_w(x)$ the final activations of a network on a sample $x$ (for example the softmax or pre-softmax vector). We assume that an attacker makes queries on $n$ images $\x = (x_1, \ldots, x_n)$, and obtains the activations $f_w(\x)$. The pipeline then is described by the Markov Chain
\[
A(\Df, \Dr) \to w \to S_{\Df}(w) \to f_{S_{\Df}(w)}(\x).
\]
The key question now is to determine how much information can an attacker recover about $\Df$, starting from the final activations $f_{S_{\Df}(w)}(\x)$? We provide a new set of bounds that quantifies the average information per query an attacker can extract from the model.

The forgetting procedure we propose is obtained using the Neural Tangent Kernel (NTK). We show that this forgetting procedure is able to handle the null space of the weights better than previous approaches when using over-parametrized models such as DNNs. In experiments, we confirm that it works uniformly better than previous proposals on all forgetting metrics introduced \cite{golatkar2019eternal}, \textit{both in the white-box and black-box case} (\Cref{fig:pca-plot}). 

Note that one may think of forgetting in a black-box setting as just changing the activations (\eg, adding noise or hiding one class output) so that less information can be extracted. This, however, is not proper forgetting as the model still contains information, it is just not visible outside. We refer to forgetting as removing information from the weights, but we provide bounds for how much information can be extracted after scrubbing in the black-box case, and show that they are order of magnitudes smaller than the corresponding bounds for white boxes for the same target accuracy.

\paragraph{Key contributions:}  To summarize, our contributions are as follow:
\begin{enumerate}
    \item We introduce methods to scrub information from, and analyze the content of deep networks from their final activations (black-box attacks).
    \item We introduce a ``one-shot'' forgetting algorithms that work better than the previous method \cite{golatkar2019eternal} for both white-box and black-box attacks for DNNs.
    \item This is possible thanks to an elegant connection between activations and weights dynamics inspired by the neural tangent kernel (NTK), which allows us to better deal with the null-space of the network weights. Unlike the NTK formalism, we do not need infinite-width DNNs. 
    \item We show that better bounds can be obtained against black-box attacks than white-box, which gives a better forgetting vs error trade-off curve.
\end{enumerate}

\section{Related work}

\paragraph{Differential privacy} \cite{dwork2014algorithmic} aims to learn the parameters of a model in such a way that no information about any particular training sample can be recovered. This is a much stronger requirement than forgetting, where we only want to remove -- \emph{after} training is done -- information about a given subset of samples. Given the stronger requirements, enforcing differential privacy is difficult for deep networks and often results in significant loss of accuracy \cite{abadi2016deep,chaudhuri2011differentially}.

\paragraph{Forgetting:}  The term ``machine unlearning'' was introduced by \cite{cao2015towards}, who shows an efficient forgetting algorithm in the restricted setting of statistical query learning, where the learning algorithm cannot access individual samples. \cite{mirzasoleiman2017deletion} provided the first framework for instantaneous data summarization with data deletion using robust streaming submodular optimization.  \cite{ginart2019making} formalizes the problem of efficient data elimination, and provides engineering principles for designing forgetting algorithms. However, they only provide a data deletion algorithms for k-means clustering. \cite{bourtoule2019machine} propose a forgetting procedure based on sharding the dataset and training multiple models. Aside from the storage cost, they  need to retrain subset of the models, while we aim for one-shot forgetting. \cite{baumhauer2020machine} proposed a forgetting method for logit-based classification models by applying linear transformation to the output logits, but do not remove information from the weights. \cite{guo2019certified} formulates data removal mechanisms using differential privacy, and provides an algorithm for convex problems based on a second order Newton update. They suggest applying this method on top of the features learned by a differentially private DNN, but do not provide a method to remove information from a DNN itself. \cite{izzo2020approximate} provides a projective residual update based method using synthetic data points to delete data points for linear regression based models. \cite{koh2017understanding,giordano2019swiss} provide a newton based method for computing the influence of a training point on the model predictions in the context of model interpretation and cross-validation, however, such an approach can also be used for data removal.  Closer to us, \cite{golatkar2019eternal} proposed a selective forgetting procedure for DNNs trained with SGD, using an information theoretic formulation and exploiting the stability of SGD \cite{hardt2015train}. They proposed a forgetting mechanism which involves a shift in weight space, and addition of noise to the weights to destroy information. They also provide an upper bound on the amount of remaining information in the weights of the network after applying the forgetting procedure. We extend this framework to activations, and show that using an NTK based scrubbing procedure uniformly improves the scrubbing procedure in all metrics that they consider.

\paragraph{Membership Inference Attacks} \cite{truex2019demystifying,hitaj2017deep,hayes2019logan,pyrgelis2017knock,shokri2015privacy,song2017machine,sablayrolles2019white} try to guess if a particular sample was used for training a model. Since a model has forgotten only if an attacker cannot guess at better than chance level, these attacks serve as a good metric for measuring the quality of forgetting. In \Cref{fig:readout-error} we construct a black-box membership inference attack similar to the shadow model training approach in \cite{shokri2015privacy}. Such methods relate to model inversion methods \cite{fredrikson2015model} which aim to gain information about the training data from the model output.

\paragraph{Neural Tangent Kernel:} \cite{jacot2018neural,lee2019wide} show that the training dynamics of a linearized version of a Deep Network --- which are described by the so called NTK matrix --- approximate increasingly better the actual training dynamics as the network width goes to infinity. \cite{arora2019exact,li2019enhanced} extend the framework to convolutional networks.
\cite{schwartz2019information} compute information-theoretic quantities using the closed form expressions for various quantities that can be derived in this settings.
While we do use the infinite width assumption, we show that the same linearization framework and solutions are a good approximation of the network dynamics during fine-tuning, and use them to compute an optimal scrubbing procedure.

\section{Out of the box forgetting}

In this section, we derive an upper-bound for how much information can be extracted by an attacker that has black-box access to the model, that is, they can query the model with an image, and obtain the corresponding output. We will then use this to design a forgetting procedure (\Cref{sec:forgetting-procedure}).

While the problem itself may seem trivial --- can the relation $A(\Df, \Dr) = w$ be inverted to extract $\Df$? --- it is made more complex by the fact that the algorithm is stochastic, and that the map may not be invertible, but still partially invertible, that is, only a subset of information about $\Df$ can be recovered. Hence, we employ a more formal information-theoretic framework, inspired by \cite{golatkar2019eternal} and that in turns generalizes Differential Privacy \cite{dwork2014algorithmic}. We provide a-posteriori bounds which provide tighter answers, and allow one to design and benchmark scrubbing procedures even for very complex models such as deep networks, for which a-priori bounds would be impossible or vacuous. In this work, we focus on a-posteriori bounds for Deep Networks and use them to design a scrubbing procedure.

\subsection{Information Theoretic formalism}
\label{sec:info-theory-formalism}

We start by modifying the framework of \cite{golatkar2019eternal}, developed for the weights of a network, to the final activations. We expect an adversary to use a readout function applied to the final activations. Given a set of images $\x=(x_0,\ldots,x_n)$, we denote by $f_w(\x) = (f(x_0), \ldots, f(x_m))$ the concatenation of their respective final activations. Let $\D_f$ be the set of training data to forget, and let $y$ be some function of $\D_f$ that an attacker wants to reconstruct (\ie, $y$ is some piece of information regarding the samples). To keep the notation uncluttered, we write $S_\Df(w) = S(w)$ for the scrubbing procedure to forget $\Df$. We then have the following Markov chain
\[
y \longleftarrow \D_f \longrightarrow w \longrightarrow S(w) \longrightarrow f_{S(w)}(\x)
\]
connecting all quantities. Using the Data Processing Inequality \cite{cover2012elements} we have the following inequalities:
\begin{equation}
\utext{ I(y; f_{S(w)}(\x)) }{Recovered information} \leq \utext{ I(\Df; f_{S(w)}(\x)) }{Black-box upper bound} \leq \utext{ I(\Df; {S(w)}) }{White box upper-bound}
\label{eq:forgetting-bounds}
\end{equation}
where $I(x; y)$ denotes the Shannon Mutual Information between random variables $x$ and $y$. Bounding the last term --- which is a general bound on how much information an attacker with full access to the weights could extract --- is the focus of \cite{golatkar2019eternal}. In this work, we also consider the case where the attacker can only access the final activations, and hence focus on the central term. As we will show, if the number of queries is bounded then the central term provides a sharper bound compared to the black-box case.

\subsection{Bound for activations}
The mutual information in the central term is difficult to compute,\footnote{Indeed, it is even difficult to define, as the quantities at play, $\Df$ and $\x$, are fixed values, but that problem has been addressed by \cite{achille2019where,achille2018emergence} and we do not consider it here.} but in our case has a simple upper-bound:
\begin{lemma}[Computable bound on mutual information] We have the following upper bound:
\begin{align}
I(\Df; f_{S(w)}(\x)) &\leq \E_{\Df}\big[\KL{ p(f_{S(w)}(\x) | \Df \cup \Dr) }{ p(f_{S_0(w)}(\x) | \Dr)}\big],
\end{align}
where $p(f_{S(w)}(\x) | \D=\Df \cup \Dr)$ is the distribution of activations after training on the complete dataset $\Df \sqcup \Dr$ and scrubbing. Similarly, $p(f_{S_0(w)}(\x) | \D=\Dr)$ is the distribution of possible activations after training only on the data to retain $\Dr$ and applying a function $S_0$ (that does not depend on $\Df$) to the weights.
\end{lemma}

The lemma introduces the important notion that we can estimate how much information we erased by comparing the activations of our model with the activations of a reference model that was trained in the same setting, but without $\Df$. Clearly, if the activations after scrubbing are identical to the activations of a model that has never seen $\Df$, they cannot contain information about $\Df$.

We now want to convert this bound in a more practical expected information gain per query. This is not yet trivial due to them stochastic dependency of $w$ on $\D$: based on the random seed $\epsilon$ used to train, we may obtain very different weights for the same dataset. Reasoning in a way similar to that used to obtain the ``local forgetting bound'' of \cite{golatkar2019eternal}, we have:
\begin{lemma}
\label{lemma:deterministic-bound}
Write a stochastic training algorithm $A$ as $A(\D, \epsilon)$, where $\epsilon$ is the random seed and $A(\D, \epsilon)$ is a deterministic function. Then,
\begin{align}
I(\Df; f_{S(w)}(\x))
\leq \E_{\Df,\epsilon}\Big[\KL{ p(f_{S(w_\D)}(\x)) }{ p(f_{S_0(w_\Dr)}(\x))}\Big]
\end{align}
where we call $w_\D = A(\D, \epsilon)$ the deterministic result of training on the dataset $\D$ using random seed $\epsilon$. The probability distribution inside the KL accounts only for the stochasticity of the scrubbing map $S(w_\D)$ and the baseline $S_0(w_\Dr)$.
\end{lemma}
The expression above is general. To gain some insight, it is useful to write it for a special case where scrubbing is performed by adding Gaussian noise.

\subsection{Close form bound for Gaussian scrubbing}

We start by considering a particular class of scrubbing functions $S(w)$ in the form
\begin{equation}
\label{eq:gaussian-scrub}
S(w) = h(w) + n,\quad n \sim N(0,\Sigma(w, \D))    
\end{equation}
where $h(w)$ is a deterministic shift (that depends on $\D$ and $w$) and $n$ is Gaussian noise with a given covariance (which may also depend on $w$ and $\D$).
We consider a baseline $S_0(w) = w + n'$ in a similar form, where $n' \sim N(0, \Sigma_0(w, \Dr))$.

Assuming that the covariance of the noise is relatively small, so that $f_{h(w)}(\x)$ is approximately linear in $w$ in a neighborhood of $h(w)$ (we drop $\D$ without loss of generality), we can easily derive the following approximation for the distribution of the final activations after scrubbing for a given random seed $\epsilon$:
\begin{equation}
f_{S(w_\D)}(\x) \sim N(f_{h(w_\D)}(\x), \nabla_{w} f_{h(w_\D)}(\x) \Sigma \nabla_{w} f_{h(w_\D)}(\x)^T) \label{eq:scrub-activations}
\end{equation}
where $\nabla_{w} f_{h(w_\D)}(\x)$ is the matrix whose row is the gradient of the activations with respect to the weights, for each sample in $\x$. Having an explicit (Gaussian) distribution for the activations, we can plug it in \Cref{lemma:deterministic-bound} and obtain:
\begin{proposition}
\label{prop:bounds}
For a Gaussian scrubbing procedure, we  have the bounds:
\begin{align}
    I(\Df; S(w)) &\leq \E_{\Df, \epsilon}\big[ \Delta w^T {\Sigma_0}^{-1} \Delta w + d(\Sigma, \Sigma_0) \big] & \text{(white-box)} \label{eq:white-box-bound}\\
    I(\Df; f_{S(w)}(\x)) &\leq \E_{\Df, \epsilon}\big[ \Delta f^T {\Sigma'_\x}^{-1} \Delta f + d(\Sigma_\x, \Sigma'_\x) \big] & \text{(black-box)} \label{eq:black-box-bound}
\end{align}
where for a $k \times k$ matrix $\Sigma$ we set $d(\Sigma, \Sigma_0) := \tr(\Sigma \Sigma_0^{-1}) + \log |\Sigma \Sigma_0^{-1}| - k$, and
\begin{align*}
    \Delta w &:= h(w_\D) - w_\Dr, \quad
    \Delta f := f_{h(w_\D)}(\x) - f_{w_\Dr}(\x) \\
    \Sigma_\x &:= \nabla_{w} f_{h(w_\D)}(\x)\, \Sigma \, \nabla_{w} f_{h(w_\D)}(\x)^T, \quad \Sigma'_\x := \nabla_{w} f_{w_\Dr}(\x) \,\Sigma_0\, \nabla_{w} f_{w_\Dr}(\x)^T.
\end{align*}
\end{proposition}

\paragraph{Avoiding curse of dimensionality:}  \Cref{prop:bounds}. Comparing \cref{eq:white-box-bound} and \cref{eq:black-box-bound}, we see that the bound in \cref{eq:white-box-bound} involves variables of the same dimension as the number of weights, while \cref{eq:black-box-bound} scales with the number of query points. Hence, for highly overparametrized models such as DNNs, we expect that the black-box bound in \cref{eq:black-box-bound} will be much smaller if the number of queries is bounded, which indeed is what we observe in the experiments (\Cref{fig:trade-off}).

\paragraph{Blessing of the null-space:} The white-box bound depends on the difference $\Delta w$ in weight space between the scrubbed model and the reference model $w_\Dr$, while the black-box bound depends on the distance $\Delta f$ in the activations. As we mentioned in \Cref{sec:introduction}, over-parametrized models such as deep networks have a large null-space of weights with similar final activations. It may hence happen that even if $\Delta w$ is large, $\Delta f$ may still be small (and hence the bound in \cref{eq:black-box-bound} tighter) as long as $\Delta w$ lives in the null-space. Indeed, we often observe this to be the case in our experiments.

\paragraph{Adversarial queries:} Finally, this should not lead us to think that whenever the activations are similar, little information can be extracted. Notice that the relevant quantity for the black box bound is $\Delta f ({J_\x} \Sigma_0 J_\x^T)^{-1} \Delta f^T$, which involves also the gradient $J_\x = \nabla_w f_{w_\Dr}(\x)$. Hence, if an attacker crafts an adversarial query $\x$ such that its gradient $J_\x$ is small, they may be able to extract a large amount of information even if the activations are close to each other. In particular, this happens if the gradient of the samples lives in the null-space of the reference model, but not in that of the scrubbed model. In \Cref{fig:trade-off} (right), we show that indeed different images can extract different amount of information.

\section{An NTK-inspired forgetting procedure}
\label{sec:forgetting-procedure}

We now introduce a new scrubbing procedure, which aims to minimize both the white-box and black-box bounds of \Cref{prop:bounds}. It relates to the one introduced in \cite{golatkar2019eternal,guo2019certified}, but it enjoys  better numerical properties and can be computed without approximations (\Cref{sec:ntk-fisher-relation}). In \Cref{sec:experiments} we show that it gives better results under all commonly used metrics.

The main intuition we exploit is that most networks commonly used are fine-tuned from pre-trained networks (\eg, on ImageNet), and that the weights do not move much during fine-tuning on $\D = \Dr \cup \Df$ will remain close to the pre-trained values. In this regime, the network activations may be approximated as a linear function of the weights. This is inspired by a growing literature on the the so called Neural Tangent Kernel, which posits that large networks during training evolve in the same way as their linear approximation \cite{lee2019wide}.  Using the linearized model we can derive an analytical expression for the optimal forgetting function, which we validate empirically. However, we observe this to be misaligned with weights actually learned by SGD, and introduce a very simple ``isoceles trapezium'' trick to realign the solutions (see Supplementary Material).

Using the same notation as \cite{lee2019wide}, we linearize the final activations around the pre-trained weights $\theta_0$ as:
\[
f_t^\lin(x) \equiv f_0(x) + \nabla_\theta f_0(x)|_{\theta=\theta_0} w_t
\]
where $w_t = \theta_t - \theta_0$ and gives the following expected training dynamics for respectively the weights and the final activations:
\begin{align}
\dot{w}_t &= - \eta \nabla_\theta f_0(\D)^T \nabla_{f^\lin_t(\D)} \mathcal{L} \\
\dot{f}^\lin_t(x) &= - \eta \Theta_0(x, \D) \nabla_{f^\lin_t(\D)} \mathcal{L}
\end{align}

The matrix $\Theta_0 = \nabla_{\theta} f(\D) \nabla_{\theta} f(\D)^T$ of size $c|\D| \times c|\D|$, where $c$ the number of classes, is called the Neural Tangent Kernel (NTK) matrix \cite{lee2019wide,jacot2018neural}. Using this dynamics, we can approximate in closed form the final training point when training with $\D$ and $\Dr$, and compute the optimal ``one-shot forgetting''  vector to jump from the weights $w_\D$ that have been obtained by training on $\D$ to the weights $w_\Dr$ that would have been obtained training on $\Dr$ alone:
\begin{proposition}
Assuming an $L_2$ regression loss,%
\footnote{This assumption is to keep the expression simple, in the Supplementary Material we show the corresponding expression for a softmax classification loss.}
the optimal scrubbing procedure under the NTK approximation is given by
\begin{equation}
\label{eq:optimal-scrubbing}
\boxed{h_\text{NTK}(w)=w + P \Gf^{T} M V}
\end{equation}
where $\Gf^{T}$ is the matrix whose columns are the gradients of the sample to forget, computed at $\theta_0$ and $w=A(\Dr \cup \Df)$.
$P = I-\Gr^{T} \Theta_{rr}^{-1}\Gr$ is a projection matrix, that projects the gradients of the samples to forget $\Gf$ onto the orthogonal space to the space spanned by the gradients of all samples to retain. The terms
$M = \big[\Theta_{ff}-\Trf^{T}\Theta_{rr}^{-1}\Trf\big]^{-1}$ and $V = [(Y_f-f_0(\Df)) + \Trf^{T}\Theta_{rr}^{-1}(Y_{r} - f_0(\Dr))]$ re-weight each direction before summing them together.
\end{proposition}
Given this result, our proposed scrubbing procedure is:
\begin{equation}
\label{eq:ntk-scrubbing}
\boxed{S_\text{NTK}(w) = h_\text{NTK}(w) + n}
\end{equation}
where $h_\text{NTK}(w)$ is as in \cref{eq:optimal-scrubbing}, and we use noise $n \sim N(0, \lambda F^{-1})$ where $F=F(w)$ is the Fisher Information Matrix computed at $h_\text{NTK}(w)$ using $\Dr$. The noise model is as in \cite{golatkar2019eternal}, and is designed to increase robustness to mistakes due to the linear approximation.

\subsection{Relation between NTK and Fisher forgetting}
\label{sec:ntk-fisher-relation}

In \cite{golatkar2019eternal} and \cite{guo2019certified}, a different forgetting approach is suggested based on either the Hessian or the Fisher Matrix at the final point: assuming that the solutions $w(\Dr)$ and $w(\D)$ of training with and without the data to forget are close and that they are both minima of their respective loss, one may compute the shift to to jump from one minimum to the other of a slightly perturbed loss landscape. The resulting ``scrubbing shift'' $w \mapsto h(w)$ relates to the newton update:
\begin{equation}
h(w) = w - H(\Dr)^{-1} \nabla_w L_\Dr(w).
\label{eq:newton-step}
\end{equation}
In the case of an $L_2$ loss, and using the NTK model, the Hessian is given by  $H(w) = \Gr^T \Gr$ which in this case also coincides with the Fisher Matrix \cite{martens2014new}. To see how this relates to the NTK matrix, consider deterimining the convergence point of the linearized NTK model, that for an $L_2$ regression is given by $w^* = w_0 + \Gr^+\, \mathcal{Y}$,
where $\Gr^+$ denotes the matrix pseudo-inverse, and $\mathcal{Y}$ denotes the regression targets. If $\Gr$ is a tall matrix (more samples in the dataset than parameters in the network), then the pseudo-inverse is $\Gr^+ = H^{-1} \Gr^{T}$, recovering the scrubbing procedure considered by \cite{golatkar2019eternal,guo2019certified}.
However, if the matrix is wide (more parameters than samples in the network, as is often the case in Deep Learning), the Hessian is not invertible, and the pseudo-inverse is instead given by $\Gr^+ = \Gr^T \Theta^{-1}$, leading to our proposed procedure. In general, when the model is over-parametrized there is a large null-space of weights that do not change the final activations or the loss. The degenerate Hessian is not informative of where the network will converge in this null-space, while the NTK matrix gives the exact point.

\section{Experiments}
\label{sec:experiments}

\subsection{Datasets}
We report experiments on smaller versions of CIFAR-10 \cite{krizhevsky2009learning} and Lacuna-10 \cite{golatkar2019eternal}, a dataset derived from the VGG-Faces \cite{Cao18} dataset. We obtain the small datasets using the following procedure: we randomly sample 500 images (100 images from each of the first 5 classes) from the training/test set of CIFAR-10 and Lacuna-10 to obtain the small-training/test respectively. We also sample 125 images from the training set (5 classes $\times$ 25 images) to get the validation set. So, in short, we have 500 (5 $\times$ 100) examples for training and testing repectively, and 125 (5 $\times$ 25) examples for validation. On both the datasets we choose to forget 25 random samples (5\% of the dataset). Without loss of generality we choose to forget samples from class 0.

\subsection{Models and Training}
We use All-CNN \cite{springenberg2014striving} (to which we add batch-normalization before non-linearity) and ResNet-18 \cite{he2016deep} as the deep neural networks for our experiments. We pre-train the models on CIFAR-100/Lacuna-100 and then fine-tune them (all the weights) on CIFAR-10/Lacuna-10. We pre-train using SGD for 30 epochs with a learning rate of 0.1, momentum 0.9 and weight decay 0.0005. Pre-training helps in improving the stability of SGD while fine-tuning. For fine-tuning we use a learning rate of 0.01 and weight decay 0.1. While applying weight decay we bias the weights with respect to the initialization. During training we always use a batch-size of 128 and fine-tune the models till zero training error. Also, during fine-tuning we do not update the running mean and variance of the batch-normalization parameters to simplify the training dynamics. We perform each experiment 3 times and report the mean and standard deviation.

\subsection{Baselines}
\begin{figure}[t]
    \centering
    \includegraphics[width=1.0\linewidth]{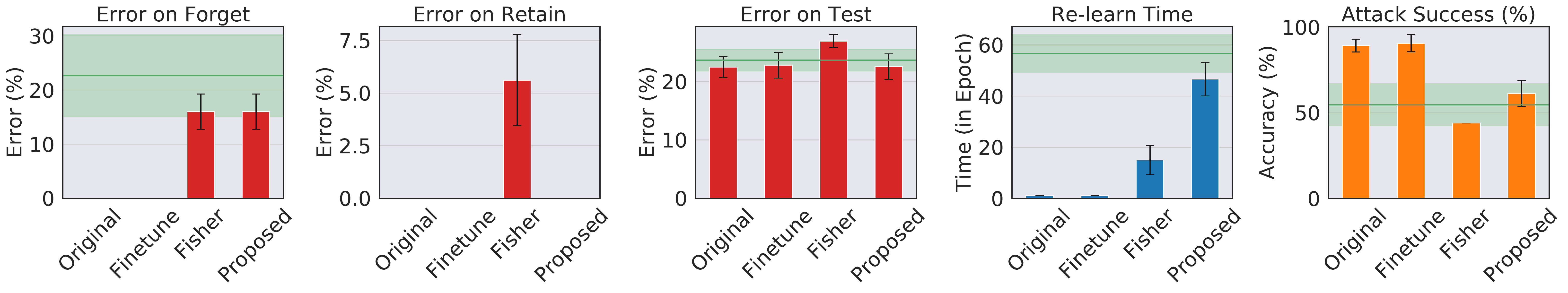}
    \includegraphics[width=1.0\linewidth]{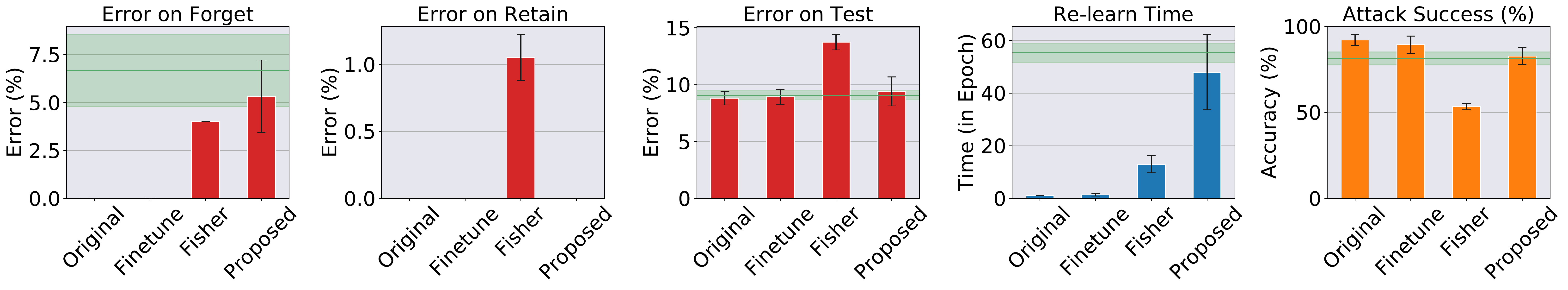}
    \caption{Comparison of different models baselines (original, finetune) and forgetting methods (Fisher \cite{golatkar2019eternal} and our NTK proposed method), using several readout functions (\textbf{(Top)} CIFAR and (\textbf{Bottom}) Lacuna). We benchmark them against a model that has never seen the data (the gold reference for forgetting): values (mean and standard deviation) measured from this models  corresponds to the green region. Optimal scrubbing procedure should lie in the green region, or they will leak information about $\Df$.
    We compute three read-out functions: \textbf{(a)} Error on forget set $\Df$, \textbf{(b)} Error on retain set $\Dr$, \textbf{(c)} Error on test set $\mathcal{D}_\text{test}$. 
    \textbf{(d)} Black-box membership inference attack: We construct a simple yet effective membership attack using the entropy of the output probabilities. We measures how often the attack model (using the activations of the scrubbed network) classify a sample belonging $\Df$ as a training sample rather than being fooled by the scrubbing.
    \textbf{(e)} Re-learn time for different scrubbing methods: How fast a scrubbed model learns the forgotten cohort when fine-tuned on the complete dataset. We measure the re-learn time as the first epoch when the loss on $\Df$ goes below a certain threshold. 
    }
    \label{fig:readout-error}
\end{figure}

We consider three baselines for comparison: (i) \textbf{Fine-tune}, we fine-tune $w(\D)$ on $\Dr$ (similar to catastrophic forgetting) and (ii) \textbf{Fisher forgetting} \cite{golatkar2019eternal}, we scrubs the weights by adding Gaussian noise using the inverse of the Fisher Information Matrix as covariance matrix, (iii) \textbf{Original} corresponds to the original model trained on the complete dataset ($w(\D)$) without any forgetting. We compare those, and our proposal, with optimal reference the model $w(\Dr)$ trained from scratch on the retain set, that is, without using $\Df$ in the first place. Values read from this reference model corresponds to the \textcolor{ao}{green} region in \Cref{fig:readout-error} and represent the gold standard for forgetting: In those plots, an optimal algorithm should lie inside the green area.

\begin{figure}[t]
    \centering
    \includegraphics[width=.99\linewidth]{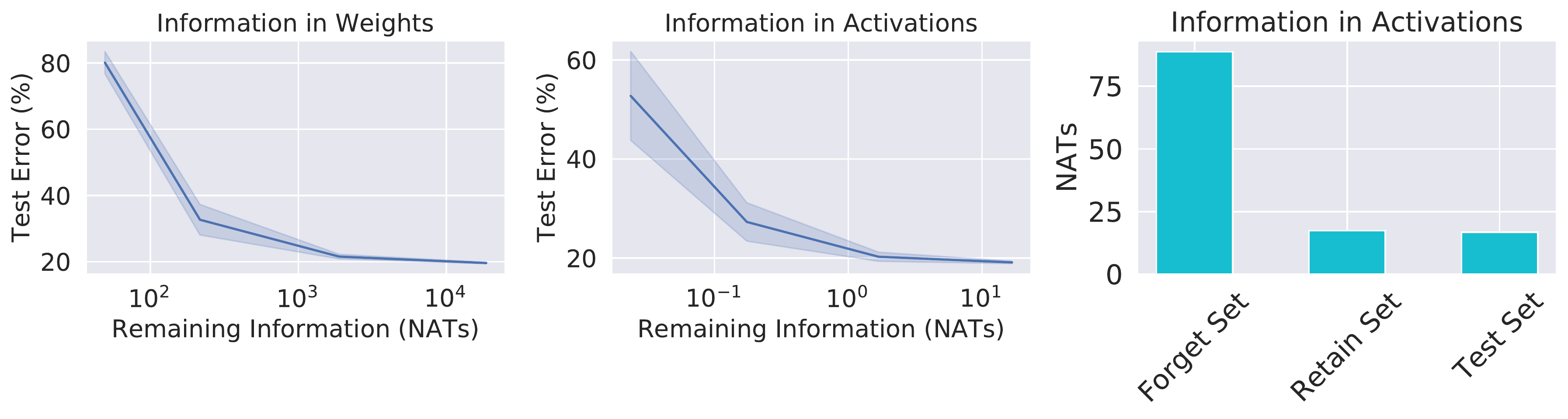}%
    \caption{
    \label{fig:trade-off}
    \textbf{Error-forgetting trade-off} Using the proposed scrubbing procedure, by changing the variance of the noise, we can reduce the remaining information in the weights \textbf{(white-box bound, left)} and activations \textbf{(black-box bound, center)}. However, it comes at the cost of increasing the test error.
    Notice that the bound on activation is much sharper than the bound on error at the same accuracy.
    \textbf{(Right) Different samples leak different information.} An attacker querying samples from $\Df$ can gain much more information than querying unrelated images. This suggest that adversarial samples may be created to leak even more information.
    } 
\end{figure}

\begin{figure}[t]
    \centering
    \includegraphics[height=3cm]{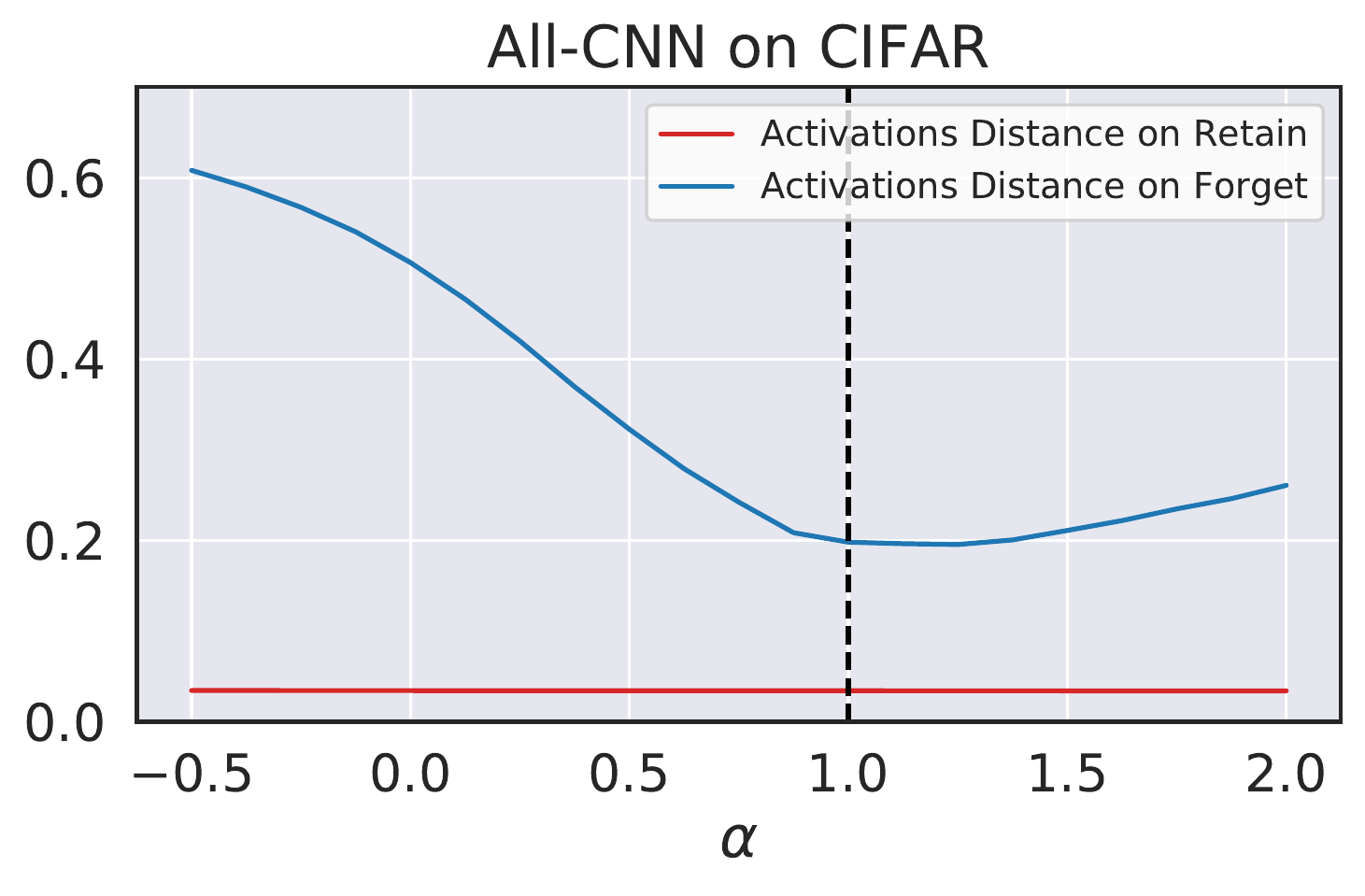}
    \hspace{.5cm}
    \includegraphics[height=3cm]{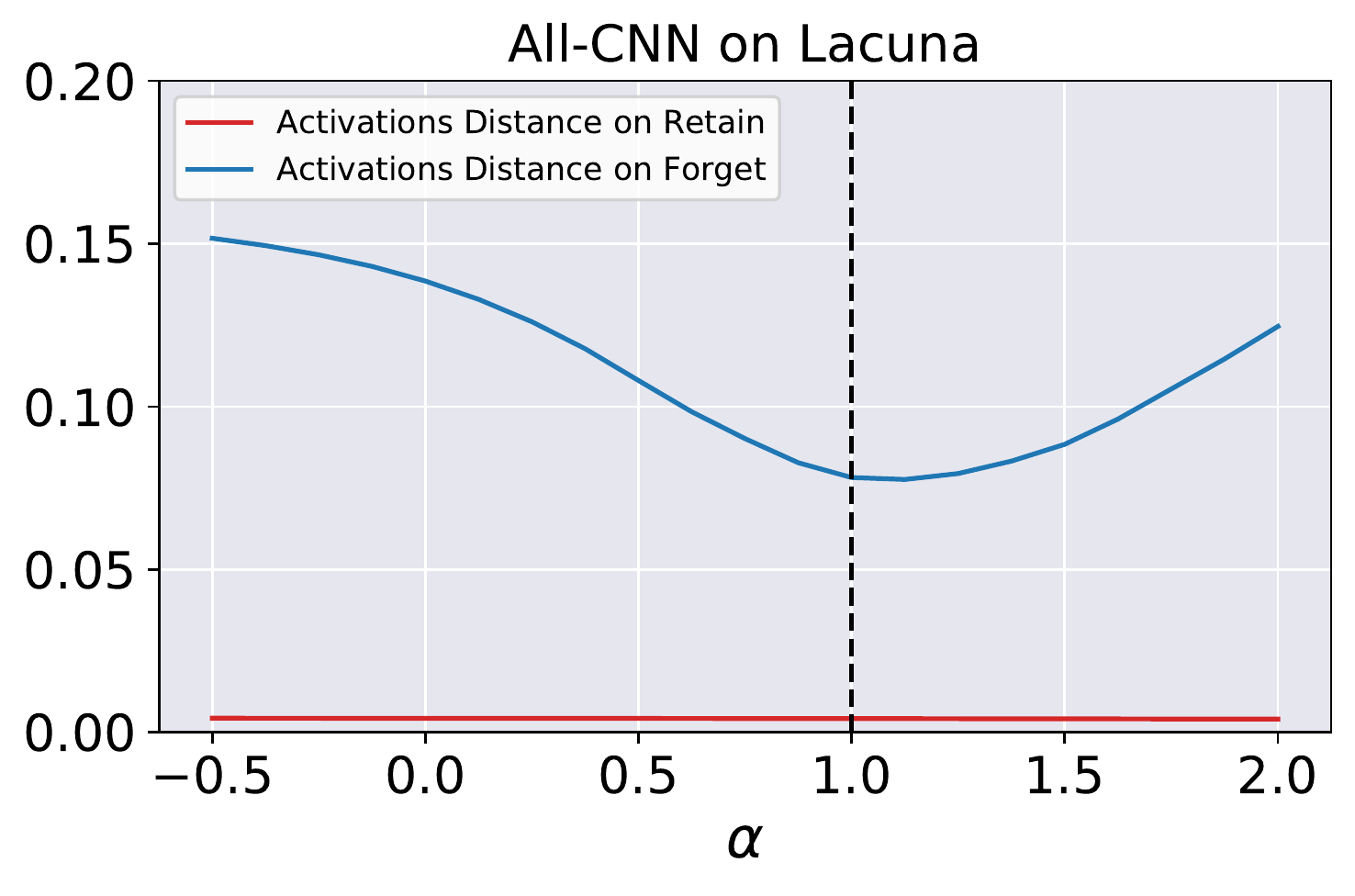}
    \caption{\textbf{Scrubbing brings activations closer to the target.} We plot the $L_1$ norm of the difference between the final activations (post-softmax) of the target model trained only on $\Dr$, and models sampled along the line joining the original model $w(D)$ ($\alpha=0$) and the proposed scrubbed model ($\alpha=1$). The distance between the activations decreases as we move along the scrubbing direction. The $L_1$ distance is already low on the retain set ($\Dr$) (red) as it corresponds to the data common to $w(\D)$ and $w(\Dr)$. However, the two models differ on the forget set ($\Df$) (blue) and we observe that the $L_1$ distance decreases as move along the proposed scrubbing direction.
    }
    \label{fig:activations-closeness}%
\end{figure}

\subsection{Readout Functions}
We use multiple readout functions similar to \cite{golatkar2019eternal}: (i) \textbf{Error on residual} (should be small), (ii) \textbf{Error on cohort to forget} (should be similar to the model re-trained from scratch on $\Dr$), (iii) \textbf{Error on test set} (should be small), (iv) \textbf{Re-learn time}, measures how quickly a scrubbed model learns the cohort to forget, when fine-tuned on the complete data. Re-learn time (measured in epochs) is the first epoch when the loss during fine-tuning (the scrubbed model) falls below a certain threshold (loss of the original model (model trained on $\D$) on $\Df$). (v) \textbf{Blackbox membership inference attack}: We construct a simple yet effective blackbox membership inference attack using the entropy of the output probabilities of the scrubbed model. Similar to the method in \cite{shokri2015privacy}, we formulate the attack as a binary classification problem (class 1 - belongs to training set and class 0 - belongs to test set). For training the attack model (Support Vector Classifier with Radial Basis Function Kernel) we use the retain set ($\Dr$) as class 1 and the test set as class 0.  We test the success of the attack on the cohort to forget ($\Df$). Ideally, the attack accuracy for an optimally scrubbed model should be the same as a model re-trained from scratch on $\Df$, having a higher value implies incorrect (or no) scrubbing, while a lower value may result in Streisand Effect, (vi) \textbf{Remaining information in the weights} \cite{golatkar2019eternal} and (vii) \textbf{Remaining information in the activations}: We compute an upper bound on the information the activations contain about the cohort to forget ($\Df$) (after scrubbing) when queried with images from different subsets of the data ($\Dr$, $\Dr$).
\label{sec:readout}

\subsection{Results}

\paragraph{Error readouts:} In \Cref{fig:readout-error} (a-c), we compare error based readout functions for different forgetting methods. Our proposed method outperforms Fisher forgetting  which incurs high error on the retain ($\Dr$) and test ($\Dtest$) set to attain the same level of forgetting. This is due the large distance between $w(\D)$ and $w(\Dr)$ in weight space, which forces it to add too much noise to erase information about $\Df$, and ends up also erasing information about the retain set $\Dr$ (high error on $\Dr$ in \Cref{fig:readout-error}). Instead, our proposed method first moves $w(\D)$ in the direction of $w(\Dr)$, thus minimizing the amount of noise to be added (\Cref{fig:pca-plot}). Fine-tuning the model on $\Dr$ (catastrophic forgetting) does not actually remove information from the weights and performs poorly on all the readout functions.

\paragraph{Relearn time:} In  \Cref{fig:readout-error} (d), we compare the re-learn time for different methods. Re-learn time can be considered as a proxy for the information remaining in the weights about the cohort to forget ($\Df$) after scrubbing. We observe that the proposed method outperforms all the baselines which is in accordance with the previous observations (in \Cref{fig:readout-error}(a-c)).

\paragraph{Membership attacks:} In \Cref{fig:readout-error} (e), we compare the robustness of different scrubbed models against blackbox membership inference attacks (attack aims to identify if the scrubbed model was ever trained on $\Df$). This can be considered as a proxy for the remaining information (about $\Df$) in the activations. We observe that attack accuracy for the proposed method lies in the optimal region (\textcolor{ao}{green}), while Fine-tune does not forget ($\Df$), Fisher forgetting may result in Streisand effect which is undesireable.

\paragraph{Closeness of activations:} In \Cref{fig:activations-closeness}, we show that the proposed scrubbing method brings the activations of the scrubbed model closer to retrain model (model retrained from scratch on $\Dr$). We measure the closeness by computing: $\mathbb{E}_{x \sim \Df/\Dr} \big[ \|f_{\text{scrubbed}}(x) - f_{\text{retrain}}(x)\|_{1} \big]$ along the scrubbing direction, where $f_{\text{scrubbed}}(x)$, $f_{\text{retrain}}(x)$ are the activations (post soft-max) of the proposed scrubbed and retrain model respectively. The distance between the activations on the cohort to forget ($\Df$) decreases as we move along the scrubbing direction and achieves a minimum value at the scrubbed model, while it almost remains constant on the retain set. Thus, the activations of the scrubbed model shows desirable behaviour on both $\Dr$ and $\Df$.

\paragraph{Error-forgetting trade-off} In \Cref{fig:trade-off}, we plot the trade-off between the test error and the remaining information in the weights and activations respectively by changing the scale of the variance of Fisher noise. We can reduce the remaining information but this comes at the cost of increasing the test error. We observe that the black-box bound on the information accessible with one query is much tighter than the white box bound at the same accuracy (compare left and center plot x-axes). Finally, in (right), we show that query samples belonging to the cohort to be forgotten ($\Df$) leaks more information about the $\Df$ rather than the retain/test set, proving that indeed carefully selected samples are more informative to an attacker than random samples.

\section{Discussion}

Recent work \cite{achille2019where,golatkar2019eternal,guo2019certified} has started providing insights on both the amount of information that can be extracted from the weights about a particular cohort of the data used for training, as well as give constructive algorithms to ``selectively forget.'' Note that forgetting alone could be trivially obtained by replacing the model with a random vector generator, obviously to the detriment of performance, or by retraining the model from scratch, to the detriment of (training time) complexity. In some cases, the data to be retained may no longer be available, so the latter may not even be an option.

We introduce a scrubbing procedure based on the NTK linearization which is designed to minimize both a white-box bound (which assumes the attacker has the weights), and a newly introduced black-box bound. The latter is a bound on the information that can be obtained about a cohort using only the observed input-output behavior of the network. This is relevant when the attacker performs a bounded number of queries. If the attacker is allowed infinitely many observations, the matter of whether the black-box and white-box attack are equivalent remains open: Can an attacker always craft sufficiently exciting inputs so that the exact values of all the weights can be inferred? An answer would be akin to a generalized ``Kalman Decomposition'' for deep networks. This alone is an interesting open problem, as it has been pointed out recently that good ``clone models'' can be created by copying the response of a black-box model on a relatively small set of exciting inputs, at least in the restricted cases where every model is fine-tuned from a common pre-trained models \cite{krishna2019thieves}. 

While our bounds are tighter that others proposed in the literature, our model has limitations. Chiefly, computational complexity.
The main source of computational complexity is computing and storing the matrix $P$ in \cref{eq:optimal-scrubbing}, which naively would require $O(c^2|\Dr|^2)$ memory and $O(|w| \cdot c^2|\Dr|^2)$ time. For this reason, we conduct experiments on a relatively small scale, sufficient to validate the theoretical results, but our method is not yet scalable to production level.
However, we notice that $P$ is the projection matrix on the orthogonal of the subspace spanned by the training samples, and there is a long history \cite{bunch1978rank} of numerical methods to incrementally compute such operator incrementally and without storing it fully in memory.
We leave these promising option to scale the method as subject of future investigation. 

\section*{Acknowledgement}
We would like to thank the anonymous reviewers for their feedback and suggestions. This work is supported by ARO W911NF-17-1-0304 and ONR N00014-19-1-2229.

\clearpage

\bibliographystyle{splncs04}
\bibliography{references}

\clearpage
\appendix

\titlerunning{Forgetting outside the box}
\authorrunning{Golatkar, Achille, Soatto}
\title{Forgetting Outside the Box\\
\normalsize{Supplementary Material}}

\author{Aditya Golatkar \and
Alessandro Achille \and
Stefano Soatto}

\institute{}

\maketitle

\setcounter{figure}{4}    

In the Supplementary Material we:
\begin{itemize}
    \item \Cref{appendix:experimental-details}: Provide implementation details for our  scrubbing procedure;
    \item \Cref{appendix:additional-experiments}: Show further experiments on more datasets (CIFAR-10, Lacuna, TinyImagenet) and models (AllCNN, ResNet).
    \item \Cref{appendix:proofs}: Provide proofs for all propositions and equations in the paper;
\end{itemize}

\begin{figure}[t]
    \centering
    \includegraphics[width=.5\linewidth]{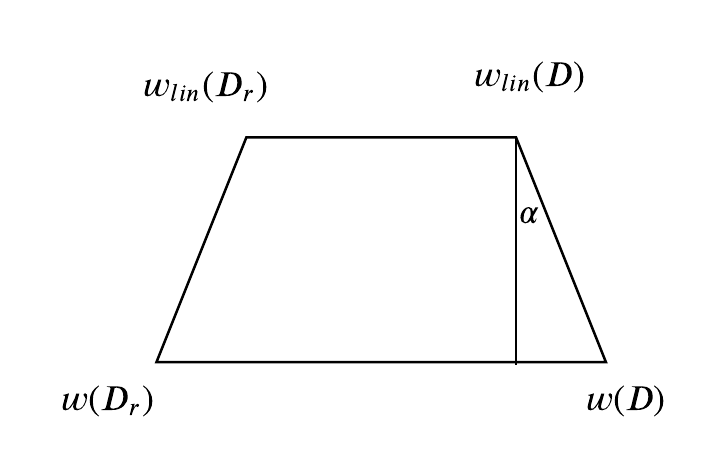}
    \caption{Isosceles Trapezium Trick: $\|w(\Dr)-w(\D)\|=\|w_\lin(\Dr)-w_\lin(\D)\|+2\sin \alpha \|w_{lin}(\D)-w(\D)\|$.
    This allows us to match outputs of the linear dynamic model with the real output, without having to match the effective learning rate of the two, and while being more robust to wrong estimation of the curvature by the linearized model.
    }
    \label{fig:trick}
\end{figure}

\section{Experimental Details}
\label{appendix:experimental-details}

We train our models with SGD (learning rate $\eta=0.01$ and momentum $m=0.9$) using weight decay ($\lambda=0.1$). All models are trained to convergence (we stop the training 5 epochs after the model achieves zero training error).

\paragraph{Pre-training and weight-decay.} In all cases, we use pre-trained models (we pre-train the models on CIFAR-100/Lacuna-100/TinyImageNet (first 150 classes) respectively), and denote by $w_0$ the pre-trained weight configuration. One important point is that we change the weight decay regularization term from the standard  $\|w\|_{2}^2$ (which pulls the weights toward zero) to $\|w-w_0\|_{2}^2$ (which pulls the weights toward the initialization). This  serves two purposes, (i) It ensures that the weights remain close to the initialization (in our case, a pre-trained network). This further helps the weight during training to remain in the neighborhood of the initialization where the the linear approximation (NTK) is good; (ii) With this change, the training dynamics of the weights/activations of a linearized network only depend on the relative change in the weights from its initial value (see \cite[Section 2.2 ]{lee2019wide} for more details).

\paragraph{NTK matrix, weight decay and cross-entropy.} For clarity, in Section~4 and Proposition~2 we only considered a unregularized MSE regression problem. To apply the theory to the more practical case of a classification cross-entropy loss with weight-decay regularization, we need the following changes.

First, using weight decay the NTK matrix becomes $\Theta=\G \G^T + \lambda I$, where $\lambda$ is the weight decay coefficient. Second, when using the cross-entropy loss, the gradients $\nabla_{f^\text{lin}_t(\D)} L$ of the loss
function can be approximated as $\nabla_{f_w(x)} L \approx \nabla_{f_{w_0(x)}} L + H_{f_{w_0}(x)}(f_w(x)-f_{w_0}(x))$, where $H_{f_{w_0}(x)}$ is the Hessian of the loss with respect to the output activations.
With these two together, we obtain $\Theta=H_{f_{w_0}(x)}\G \G^T + \lambda I$ as the NTK matrix to use in our setting. 

We did however notice that replacing $H_{f_{w_0}(x)}$ with the identity matrix $I$ works better in practice. This may be due to $H_{f_{w_0}(x)}$ estimating the wrong curvature when the softmax saturates.
Also we found that --- while in principle identical as long as the network remains in the linear regime --- linearizing around $w(\D)$ provides a better estimate of the scrubbing direction compared to linearizing around $w_0$.

\paragraph{Trapezium trick.} We observe that the linearized dynamics of in eq.~(8) and eq.~(9) correctly approximate the training direction but they usually undershoot and give a smaller norm solution than SGD. This may be due to difficulty in matching the learning rate of continuous gradient descent and discrete SGD.
To overstep these issues in a robust way, we use the following simple ``trapezium trick'' (\Cref{fig:trick}) to renormalize the scrubbing vector obtained with the linear dynamics: Instead of trying to predicting the unknown $w(\Dr)$ directly using the scrubbing vector suggested by \cref{eq:optimal-scrubbing}, we compute the two final points of the linearized dynamics $w_{\text{lin}}(\D)$ and $w_{\text{lin}}(\Dr)$, and approximate $w(\Dr)$ by constructing the isosceles trapezium in \Cref{fig:trick}. Effectively, this rescales the ideal linearized forgetting direction $w_{\text{lin}}(\Dr) - w_{\text{lin}}(\D)$ to correct for the undershooting.

\section{Additional Experiments}
\label{appendix:additional-experiments}

\begin{figure}
    \centering
    \includegraphics[height=4.0cm]{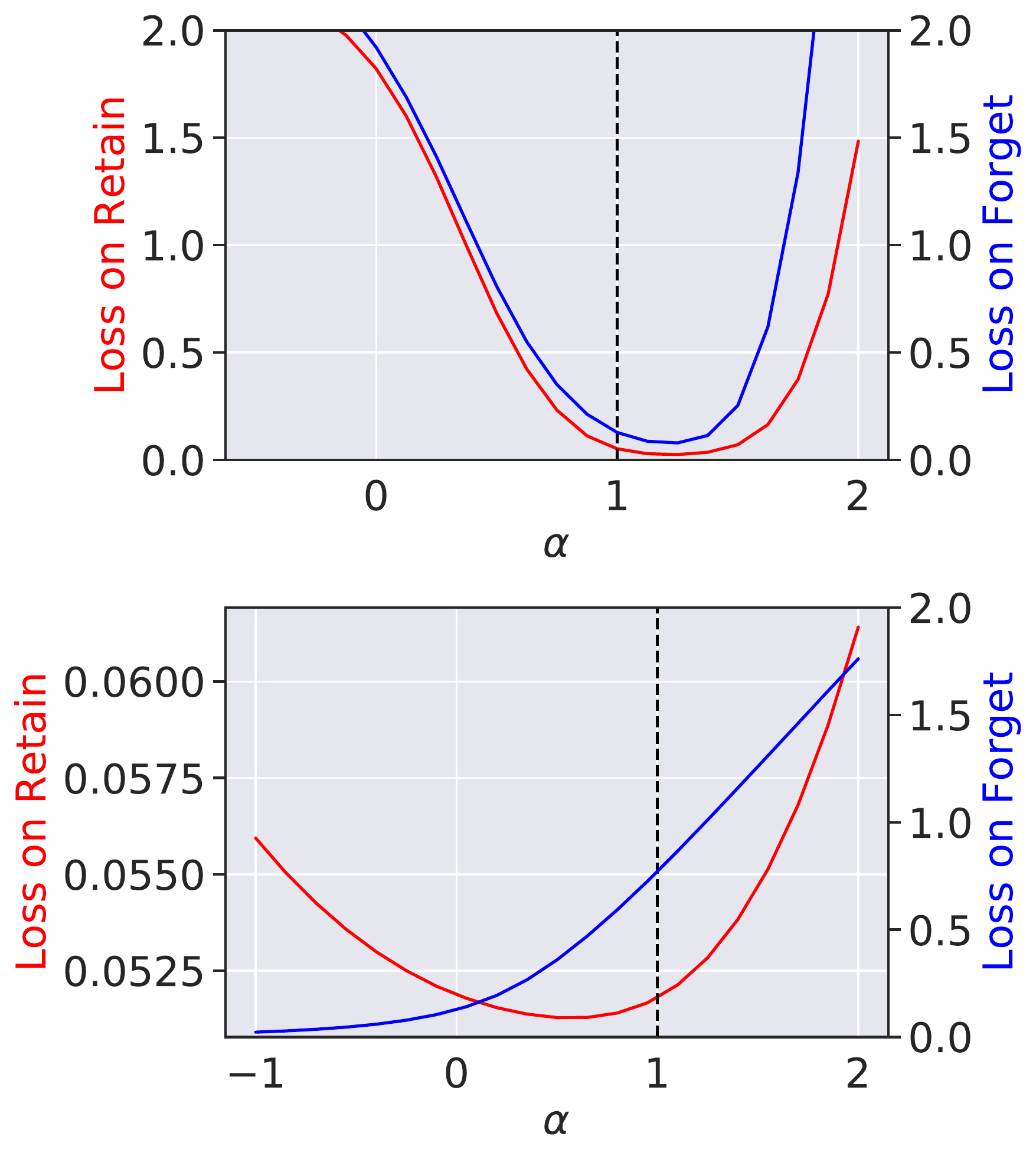} \hspace{.4cm}
    \includegraphics[height=4.0cm]{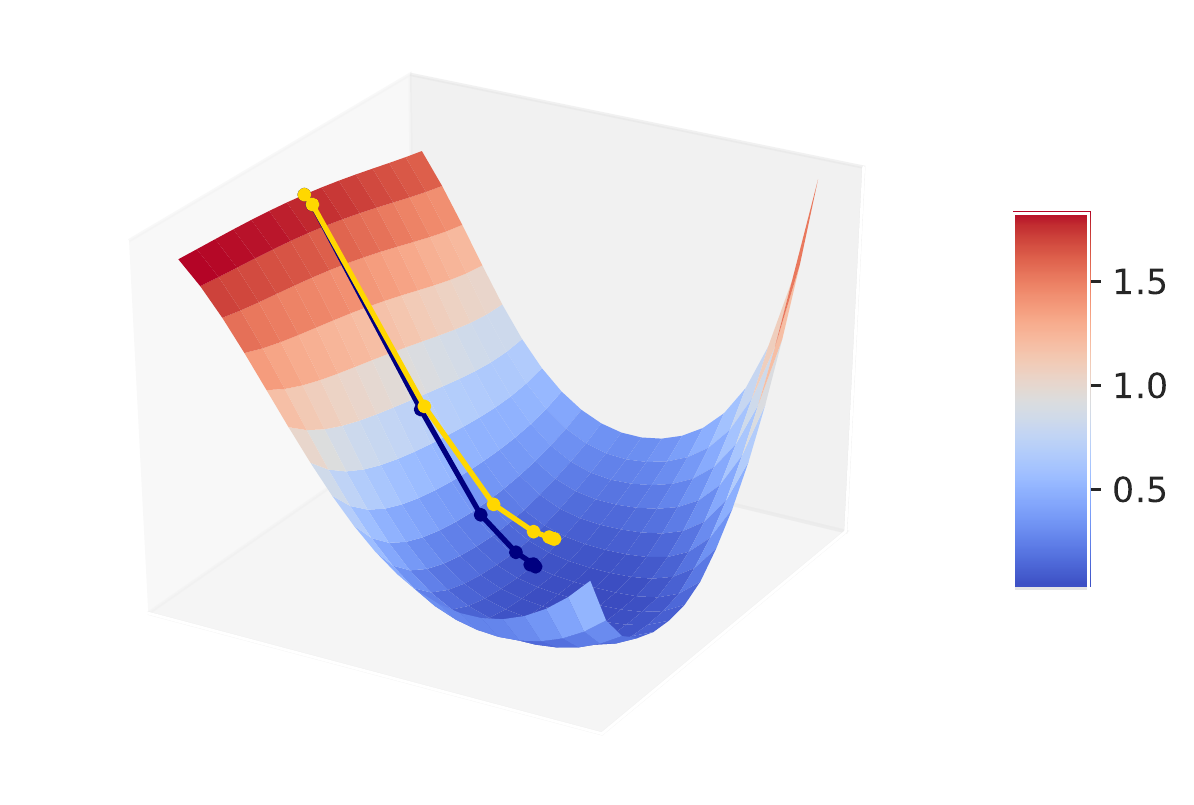}%
    \caption{
    \textbf{(Right) The loss landscape and training dynamics after pretraining are smooth and regular.}  This justifies our linearization approach to study the dynamics. The black and yellow lines are the training paths on $\D$ and $\Dr$ respectively. Notice that they remain close. \textbf{(Upper left) Loss along the line} joining the model at initialization ($\alpha=0$) and the model after training on $\D$ ($\alpha=1$) (the black path). \textbf{(Lower left)} Loss along the line joining the end point of the two paths ($\alpha=0$ and 1 respectively), which is the ideal scrubbing direction.
    \label{fig:loss-landscape}%
    }
\end{figure}

In \Cref{fig:loss-landscape} we use a PCA projection to show the geometry of the loss landscape after convergence and the training paths obtained by training the weights on the full dataset $\D$ or only on $\Dr$. We observe that the loss landscape around the pretraining point is smooth and almost convex. Moreover, the two training paths remain close to each other. This supports our choice of using a simple linearized approximation to compute the shift that jumps from one path to the other.

\begin{figure}[h]
    \centering
    \includegraphics[width=1.0\linewidth]{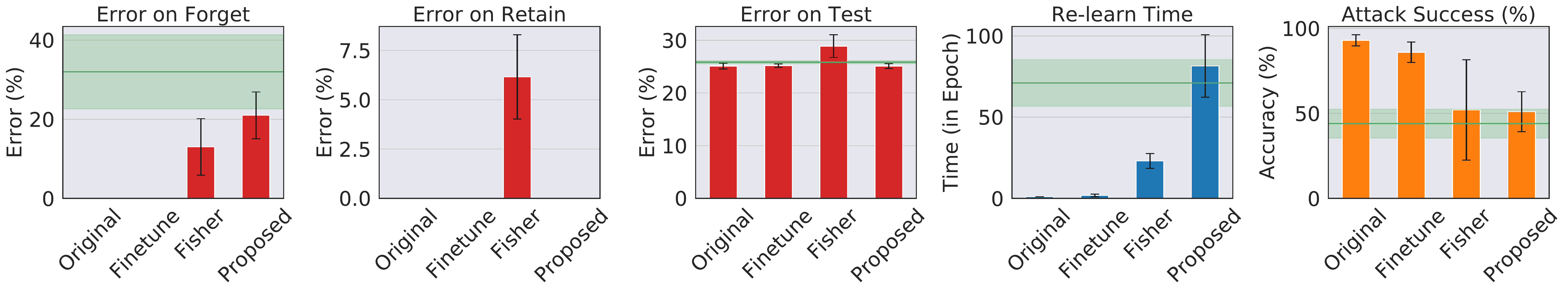}
    \includegraphics[width=1.0\linewidth]{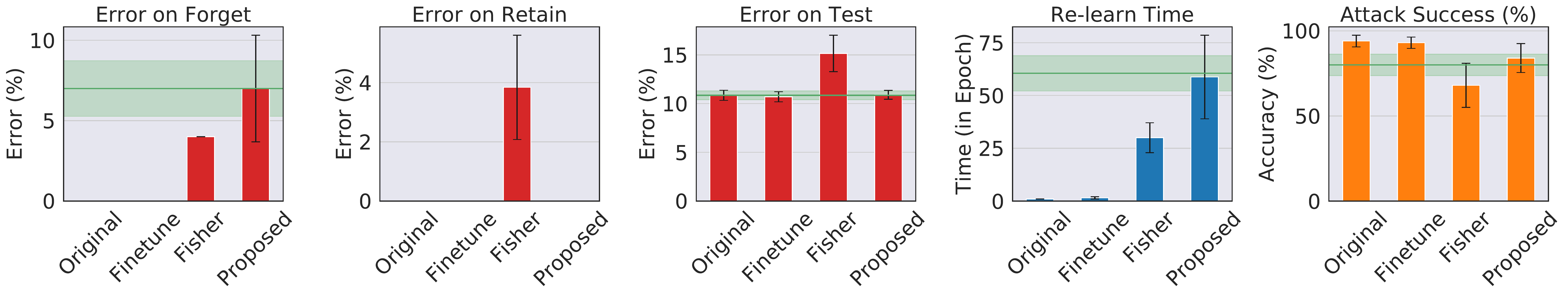}
    \includegraphics[width=1.0\linewidth]{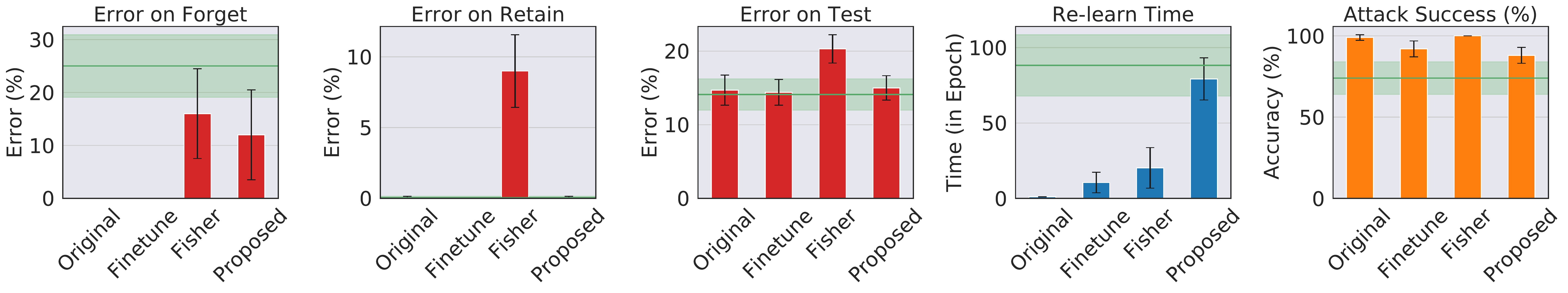}
    \includegraphics[width=1.0\linewidth]{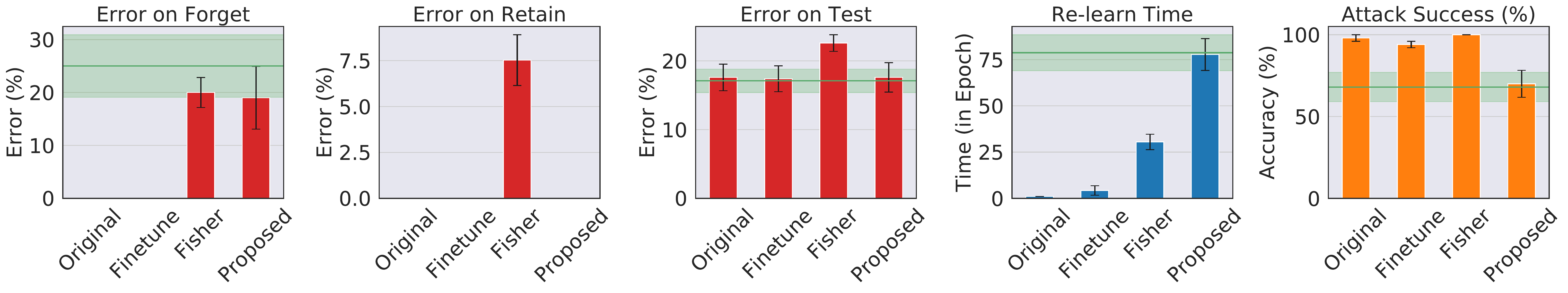}
    
    \caption{Same experiment as \Cref{fig:readout-error} for different architectures and datasets. \textbf{(Row 1)}: ResNet-18 on CIFAR, 
    \textbf{(Row 2)}: ResNet-18 on Lacuna, 
    \textbf{(Row 3)}: All-CNN on TinyImageNet and \textbf{(Row 4)}: ResNet-18 on TinyImageNet. 
    In all the experiments we observe that for different readout functions the proposed method lies in the green (target) region.  
    }
    \label{fig:readout-error-mult}
\end{figure}

\begin{figure}
    \centering
    \includegraphics[width=.99\linewidth]{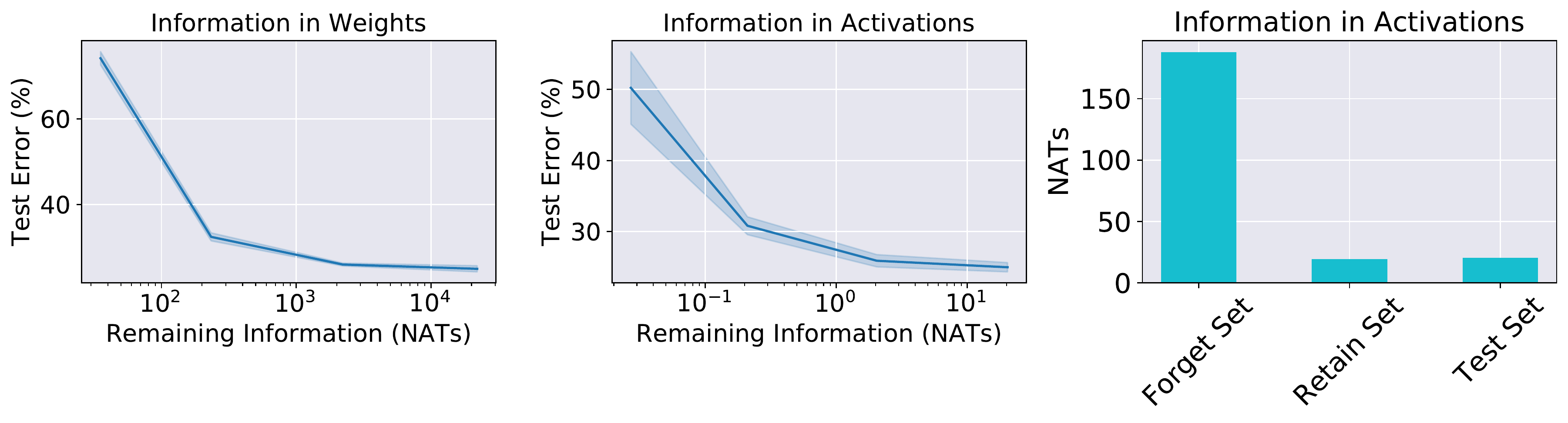}
    \includegraphics[width=.99\linewidth]{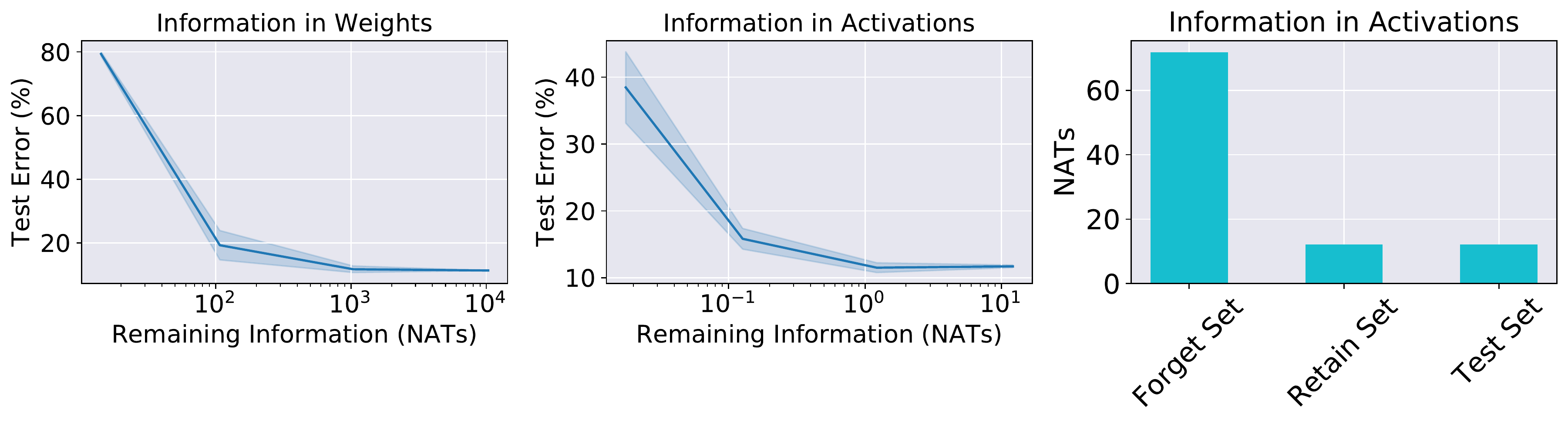}
    \includegraphics[width=.99\linewidth]{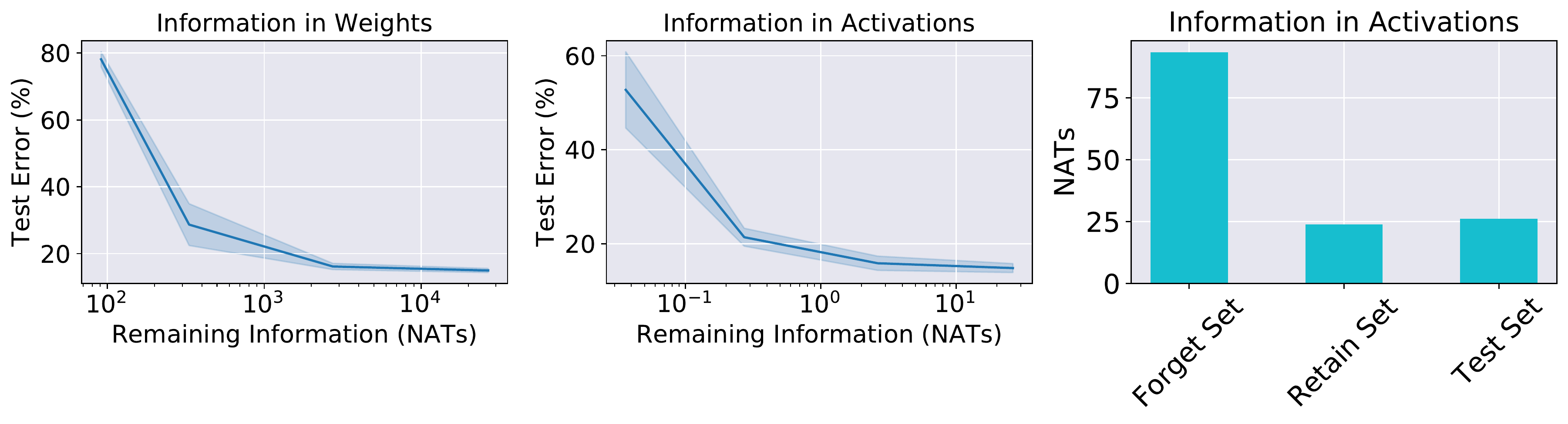}
    \includegraphics[width=.99\linewidth]{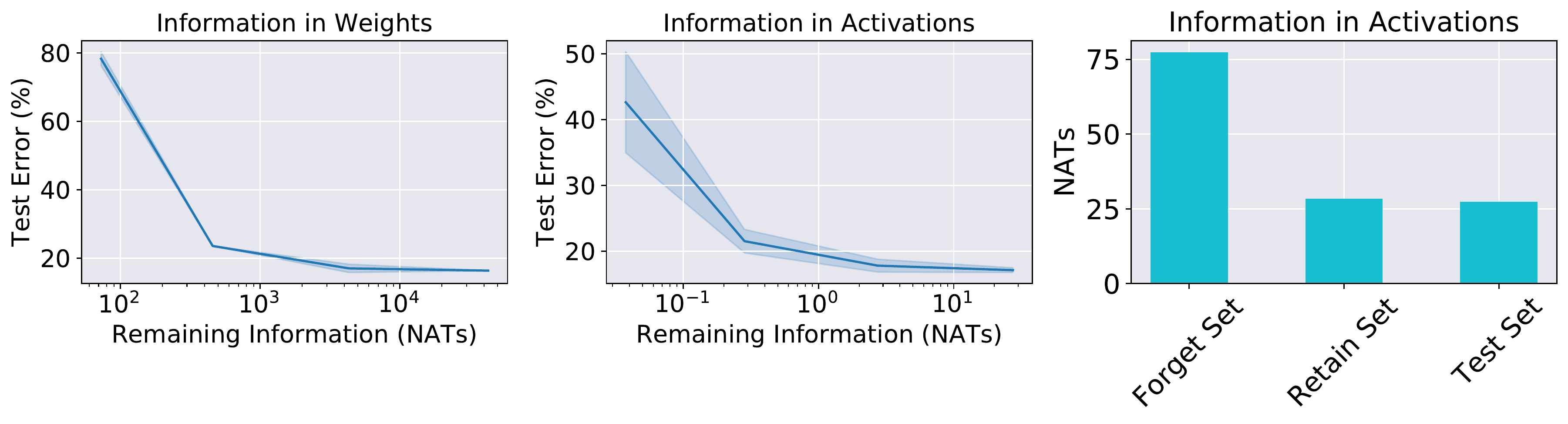}
    \caption{
    Same experiment as \Cref{fig:trade-off} for different architectures and datasets. \textbf{(Row-1)}: ResNet-18 on CIFAR, \textbf{(Row-2)}: ResNet-18 on Lacuna, \textbf{(Row-3)}: All-CNN on TinyImageNet, \textbf{(Row-4)}: ResNet-18 on TinyImageNet.
    We observe consistent behaviour across different architectures and datasets.
    } 
\label{fig:trade-off-mult}
\end{figure}

In \Cref{fig:readout-error-mult} and \Cref{fig:trade-off-mult} we show additional experiments on several architectures (ResNet-18, All-CNN) and datasets (Lacuna, CIFAR-10, TinyImageNet). In all cases we observe a similar qualitative behavior to the one discussed in the paper.

\section{Proofs}
\label{appendix:proofs}

\paragraph{Markov chain in \Cref{sec:info-theory-formalism}.}
We consider the retain set $\Dr$ (not shown) as an observed random variable, while the cohort to forget $\Df$ is an hidden variable sampled randomly from the data distribution. The directed edge $\Df \rightarrow w$ in the Markov chain derives from the fact that we first sample  $\Df$, to obtain the full complete training set $\D = \Dr \sqcup \Df$, and then train the network on $\D$ to obtain the weights $w$.

\paragraph{Proof of Lemma 1.}
We have the following upper-bound for $I(\Df; f_{S(w)}(\x))$:
\begin{align*}
I(\Df; f_{S(w)}(\x))
&= I(\Df; f_{S(w)}(\x)) \\
&= \E_{\Df}\big[\KL{ p(f_{S(w)}(\x) | \Df \cup \Dr) }{ p(f_{S(w)}(\x)| \Dr)}\big]\\
&= \E_\Df \E_{p(f_{S(w)}(\x) | \Df \cup \Dr)}\Big[ \log \frac{p(f_{S(w)}(\x) | \Df \cup \Dr)}{ p(f_{S(w)}(\x)|\Dr)}\Big]\\
&= \E_\Df \E_{p(f_{S(w)}(\x) | \Df \cup \Dr)}\Big[  \log \frac{p(f_{S(w)}(\x) | \Df \cup \Dr)}{ p(f_{S(w)}(\x)|\Dr)}
\\
&\hspace{5cm}+
\log \frac{p(f_{S_0(w)}(\x)|\Dr)}{ p(f_{S_0(w)}(\x)|\Dr)}
\Big]\\
&=\E_{\Df}\big[\KL{ p(f_{S(w)}(\x) | \Df \cup \Dr)}{ p(f_{S_0(w)}(\x) | \Dr)}\big]\\
&\hspace{3cm}
- \KL{p(f_{S(w)}(\x)|\Dr)}{p(f_{S_0(w)}(\x) | \Dr)}\\
&\leq \E_{\Df}\big[\KL{ p(f_{S(w)}(\x) | \Df \cup \Dr) }{ p(f_{S_0(w)}(\x) | \Dr)}\big]
\end{align*}
where the last inequality follows from the fact that KL-divergence is always non-negative.

\paragraph{Proof of Lemma 2.}
We have the inequalities:
\begin{align*}
I(\Df; f_{S(w)}(\x)) &\leq \E_{\Df}\big[\KL{ p(f_{S(w_\D)}(\x)) }{ p(f_{S_0(w_\Dr)}(\x))}\big]\\
&\leq \E_{\Df,\epsilon}\Big[\KL{ p(f_{S(w_\D)}(\x)) }{ p(f_{S_0(w_\Dr)}(\x))}\Big]
\end{align*}
where the first inequality comes from Lemma 1, and the second inequality is from \cite[Proposition 2]{golatkar2019eternal}.

\paragraph{Proof of eq. (5).}
The activations of a scrubbed network (using the Gaussian scrubbing procedure in \cref{eq:gaussian-scrub}) for a given sample $\x$ are given by $f_{h(w) + n}(\x)$, where $n \sim N(0,\Sigma)$. By linearizing the activations (using NTK formalism) around $h(w)$, we obtain the distribution of the scrubbed activations: 
\begin{equation*}
f_{h(w)+n}(\x) \sim N(f_{h(w)}(\x), \nabla_{w} f_{h(w)}(\x) \Sigma \nabla_{w} f_{h(w)}(\x)^T)
\end{equation*}
For the original model we compute this at $w=w_\D$ and take $h$ to be the the NTK scrubbing shift in \cref{eq:optimal-scrubbing}. The baseline model does not use any shift, so $h(w)=w$, and is computed at $w = w_\Dr$.

\subsubsection*{Proof of Proposition 1, \cref{eq:white-box-bound}.} As in \cite[Example 2]{golatkar2019eternal}.

\subsubsection*{Proof of Proposition 1,  \cref{eq:black-box-bound}.}
Using \Cref{eq:scrub-activations} we write the distribution of the activations of a scrubbed network:
\begin{equation*}
f_{S(w_\D)}(\x) \sim N(f_{h(w_\D)}(\x), J \Sigma J^T)
\end{equation*}
where $J=\nabla_{w} f_{h(w_\D)}(\x)$. We can similarly write the activations for baseline as:
\begin{equation*}
f_{S_0(w_\Dr)}(\x) \sim N(f_{w_\Dr}(\x), J' \Sigma_0 J'^T)
\end{equation*}
where $J'=\nabla_{w} f_{w_\Dr}(\x)$
Using the two distributions, we rewrite the bound in Lemma 2 as:
\begin{align*}
I(\Df; f_{S(w)}(\x)) &\leq \E_{\Df, \epsilon}\big[\KL{ N(f_{h(w_\D)}(\x), J \Sigma J^T)}{ N(f_{w_\Dr}(\x), J' \Sigma_0 {J'}^T}\big] \\
&= \E_{\Df, \epsilon}\big[
\Delta f^T {\Sigma'_\x}^{-1} \Delta f + 
\tr( \Sigma_\x{\Sigma_\x'}^{-1}) - \log |\Sigma_\x {\Sigma_\x'}^{-1}|  - n
\big]
\end{align*}
where $\Delta f = f_{h(w_\D)}(\x) - f_{w_\Dr}(\x)$, $\Sigma_\x=J \Sigma J^T$, $\Sigma_\x' = J' \Sigma_0 {J'}^T$, and we used the closed form expression for the KL divergence of two normal distributions.

\subsubsection*{Proof of Proposition 2.}

Let $\D=\Dr\cup\Df$ be the complete training set. To keep the notation simpler, we assume that the loss is an mean-square-error regression loss (we discuss classification using cross-entropy in Appendix~B). 

Let $w_0$ be the weights obtained after pre-training on $\mathcal{D}_\text{pre-train}$. Taking inspiration from the NTK analysis \cite{jacot2018neural,lee2019wide} we approximate the activations $f_w(x)$ for a test datum $x$ after fine-tuning on $\D$ using the liner approximation $f^\lin_w(x) = f_{w_0}(x) + \nabla_w f_{w_0}(x)(w-w_0)$. To keep the notation uncluttered, we write $f_0(x)$ instead of $f_{w_0}(x)$.

The training dynamics under the linear approximation (assuming continuous gradient descent) are then given by eq.~(8) and (9), and will converge at the final solution (see \cite{lee2019wide} for more details): \[w_\lin(\mathcal{\D}) = \G^T\Theta^{-1}(f_0(\D)-Y) + w_0.\]
Here $\G \in \mathbb{R}^{Nc \times p}$  is the gradient of the output with respect to the parameters (at initialization) for all the samples in $\D$ stacked along the rows to form a matrix ($p$ is the number of parameters in the model, $N=|\D|$ and $c$ is the number of classes),  $\Theta=\G \G^T\ (\in \mathbb{R}^{Nc \times Nc})$ is the NTK matrix. Similarly, $Y$ is the matrix formed by stacking all ground-truth labels one below the other and $f_0(\D) \in \mathbb{R}^{(Nc \times 1)}$ are the stacked outputs of the DNN at initialization on $\D$.

Similarly the baseline solution (training only on the data to retain) is:
\[w_\lin(\Dr) = \Gr^T\Theta^{-1}_{rr}(f_0(\Dr)-Y_r)+w_0,\]
where $\Trr=\Gr^{T}\Gr$.
The optimal scrubbing vector (at least for the linearized model) would then be $\delta w:= w_\lin(\Dr) - w_\lin(\D)$: adding $\delta w$ to the weights obtained by training on $\D$ makes us forget the extra examples ($\Df$), so that we obtain weights equivalent to training on $\Dr$ alone. We now derive the simplified expression in \cref{eq:optimal-scrubbing} for the optimal scrubbing vector $\delta w$.
We start by rewriting $w_\lin(\mathcal{\D})$ using block matrixes:
\begin{align*}
    w_\lin(\mathcal{\D}) &= \G^T\Theta^{-1}(f_0(\D)-Y)+w_0\\
    &= \Big[\nabla f_0(\Dr)^T \nabla f_0(\Df)^T\Big]
    {\begin{bmatrix}
    \Theta_{rr}&\Theta_{rf}\\
    \Theta_{rf}^{T}&\Theta_{ff}
    \end{bmatrix}}^{-1}
    \begin{bmatrix}
    f_0(\Dr)-Y_r\\
    f_0(\Df)-Y_f
    \end{bmatrix}+w_0
\end{align*}
We can expand the inverse of the NTK matrix using the following equations:
\begin{align*}
    {\begin{bmatrix}
    \Theta_{rr}&\Theta_{rf}\\
    \Theta_{rf}^{T}&\Theta_{ff}
    \end{bmatrix}}^{-1} &=
    \begin{bmatrix}
    \Big[\Trr-\Trf\Tff^{-1}\Trf^{T}\Big]^{-1}& -\Trr^{-1}\Trf M\\
    -M\Trf^{T}\Trr^{-1} & M
    \end{bmatrix}
\end{align*}
Where $M=\big[\Theta_{ff}-\Trf^{T}\Theta_{rr}^{-1}\Trf\big]^{-1}$. Using Woodbury Matrix Identity: \[(A+UCV)^{-1}=A^{-1}-A^{-1}U(C^{-1}+VA^{-1}U)^{-1}VA^{-1}\]
We obtain:
\[\Big[\Trr-\Trf\Tff^{-1}\Trf^{T}\Big]^{-1}=\Trr^{-1}+\Trr^{-1}\Trf M \Trf^{T}\Trr^{-1}\]
Thus,
\begin{align*}
    {\begin{bmatrix}
    \Theta_{rr}&\Theta_{rf}\\
    \Theta_{rf}^{T}&\Theta_{ff}
    \end{bmatrix}}^{-1} &=
    \begin{bmatrix}
    \Trr^{-1}+\Trr^{-1}\Trf M \Trf^{T}\Trr^{-1}&\  \ -\Trr^{-1}\Trf M\\
    -M\Trf^{T}\Trr^{-1} & M
    \end{bmatrix}
\end{align*}
Using the above relation we get
\begin{align*}
    w_\lin(\D)&=
    \Gr^T\Big(\Trr^{-1}+\Trr^{-1}\Trf M \Trf^{T}\Trr^{-1}\Big)(f_0(\Dr)-Y_r)\\
    &-\Gf^T M \Trf^{T}\Trr^{-1}(f_0(\Dr)-Y_r)\\
    &-\Gr^T\Trr^{-1}\Trf M (f_0(\Df)-Y_f)\\
    &+\Gf^T M (f_0(\Df)-Y_f) + w_0.
\end{align*}
Finally, using this the optimal shift $\delta w$ for scrubbing the weights is
\begin{align*}
    \Delta w &= w_\lin(\Dr)-w_\lin(D)\\
    &=-\Gr^T\Trr^{-1}\Trf M \Trf^{T}\Trr^{-1}(f_0(\Dr)-Y_r)\\
    &+\Gf^T M \Trf^{T}\Trr^{-1}(f_0(\Dr)-Y_r)\\
    &+\Gr^T\Trr^{-1}\Trf M (f_0(\Df)-Y_f)\\
    &-\Gf^T M (f_0(\Df)-Y_f)\\
    &=\Big[I-\Gr^T\Trr^{-1}\Gr\Big]\Gf^T M \Big[\Trf^{T}\Trr^{-1}(f_0(\Dr)-Y_r)\Big]\\
    &+\Big[I-\Gr^T\Trr^{-1}\Gr\Big]\Gf^T M \Big[(Y_f-f_0(\Df))\Big]\\
    &=P\Gf^T M V,
\end{align*}
where $P = I-\Gr^{T} \Theta_{rr}^{-1}\Gr$,
$M = \big[\Theta_{ff}-\Trf^{T}\Theta_{rr}^{-1}\Trf\big]^{-1}$ and $V = [(Y_f-f_0(\Df)) + \Trf^{T}\Theta_{rr}^{-1}(Y_{r} - f_0(\Dr))]$.

\end{document}